\documentclass{bmvc2k}

\usepackage{xcolor}
\usepackage{amsfonts}
\usepackage{bigstrut}
\usepackage{multirow}
\usepackage{booktabs}
\usepackage{bbm}
\usepackage{subfig}
\usepackage{algorithm}
\usepackage{algpseudocode}
\usepackage{bbding}

\usepackage{newtxtext,newtxmath}

\usepackage{wrapfig}
\usepackage{colortbl}

\usepackage{enumitem}
\setitemize[1]{topsep=0.3em, itemsep=-0.3em}

\title{Open-Vocabulary Object Detection with \\ Meta Prompt Representation and \\ Instance Contrastive Optimization}

\addauthor{Zhao Wang}{zwang21@cse.cuhk.edu.hk}{1}
\addauthor{Aoxue Li}{lax@pku.edu.cn}{2}
\addauthor{Fengwei Zhou}{fzhou@connect.ust.hk}{2}
\addauthor{Zhenguo Li}{li.zhenguo@huawei.com}{2}
\addauthor{Qi Dou}{qidou@cuhk.edu.hk}{1}

\addinstitution{
 Department of Computer Science and Engineering\\
 The Chinese University of Hong Kong\\
 Shatin, Hong Kong
}
\addinstitution{
 Huawei, China
}
\runninghead{Wang et al}{Open-Vocabulary Object Detection with MIC}

\def\eg{\emph{e.g}\bmvaOneDot}

\def\etal{\emph{et al}\bmvaOneDot}

\begin{document}

\maketitle

\begin{abstract}
Classical object detectors are incapable of detecting novel class objects that are not encountered before. Regarding this issue, Open-Vocabulary Object Detection (OVOD) is proposed, which aims to detect the objects in the candidate class list. However, current OVOD models are suffering from overfitting on the base classes, heavily relying on the large-scale extra data, and complex training process. To overcome these issues, we propose a novel framework with Meta prompt and Instance Contrastive learning (MIC) schemes. Firstly, we simulate a novel-class-emerging scenario to help the prompt learner that learns class and background prompts generalize to novel classes. Secondly, we design an instance-level contrastive strategy to promote intra-class compactness and inter-class separation, which benefits generalization of the detector to novel class objects. Without using knowledge distillation, ensemble model or extra training data during detector training, our proposed MIC outperforms previous SOTA methods trained with these complex techniques on LVIS. Most importantly, MIC shows great generalization ability on novel classes, \eg, with $+4.3\%$ and $+1.9\% \ \mathrm{AP}$ improvement compared with previous SOTA on COCO and Objects365, respectively.

\end{abstract}

\section{Introduction}

\begin{figure}[t]%
    \centering
    \subfloat[]{
        \includegraphics[width=0.73\columnwidth]{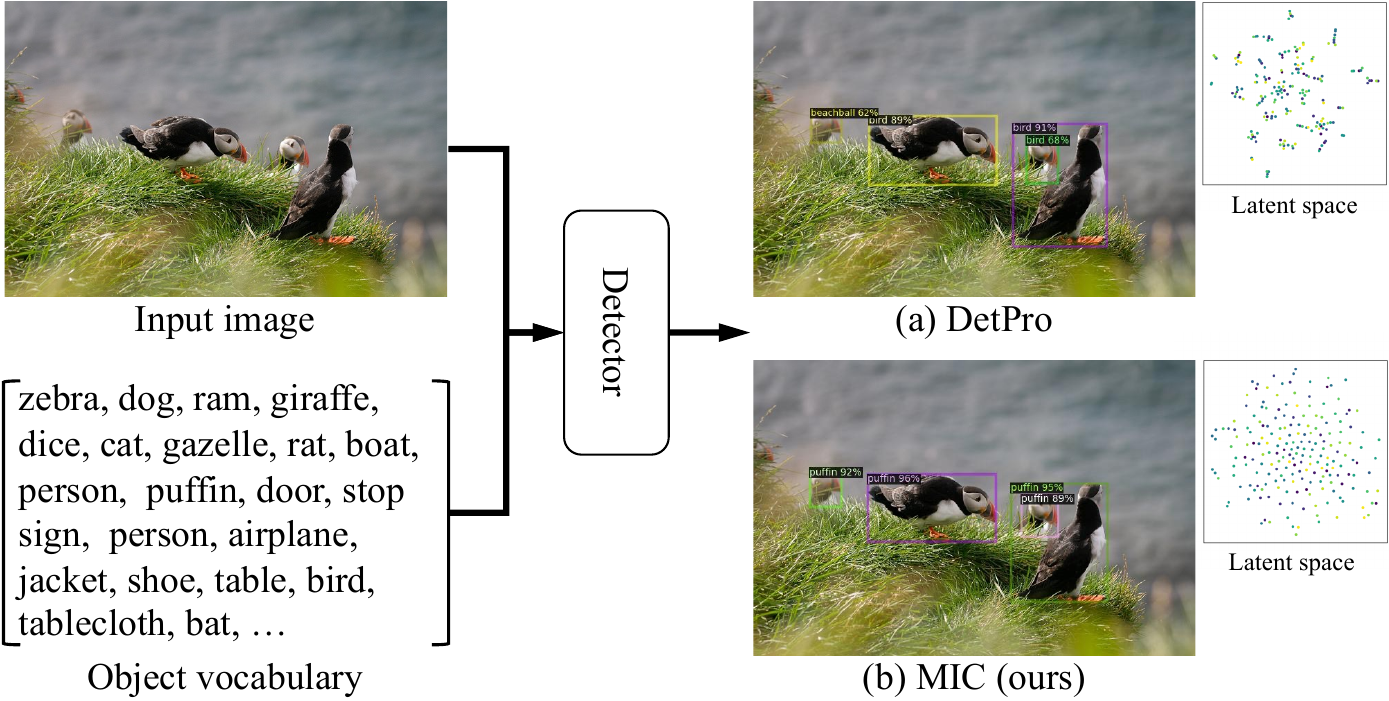}
        \label{fig:teaser}
        }
        % \hfill
    \subfloat[]{
        \includegraphics[width=0.23\columnwidth]{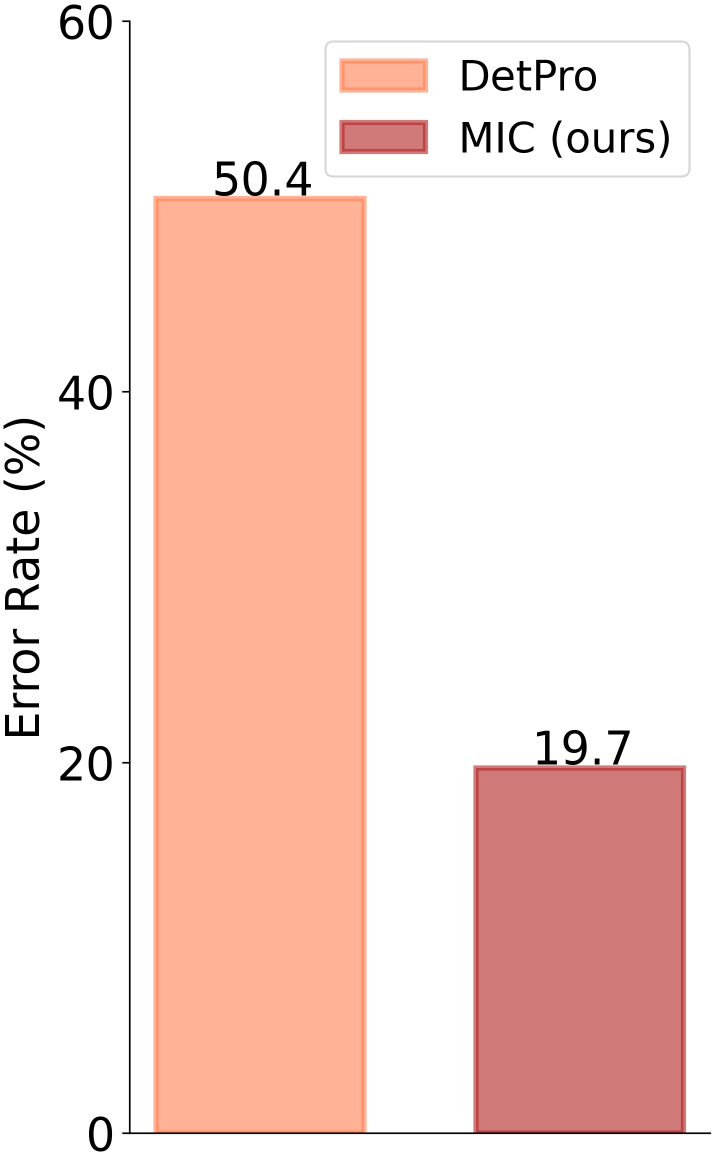}
        \label{fig:novel_error}
        }
    \vspace{2mm}
    \caption{(a) In OVOD, the detector aims to detect any objects within an object vocabulary in an input image. Previous method, \eg, DetPro, can easily misclassify some highly similar classes (puffin v.s. bird). Our method improves the model generalization ability, which can be more discriminative to these similar categories. Note that every point indicates a category in the latent space; (b) The error rate of predicting novel objects as base ones.}
\end{figure}

Deep learning models have been successful in closed-set large-scale object detection, in which the carefully designed detectors \cite{ren2015faster,he2017mask,carion2020end,zhu2021deformable} can accurately localize and classify the objects learned from the training set.
However, these classical detectors always fail to generalize to unseen novel class objects during inference.
Thanks to recent advancements in vision-language models \cite{radford2021learning,jia2021scaling,li2022grounded}, ViLD \cite{gu2022openvocabulary} extends the traditional closed-set object detection to an open-set scenario, named Open-Vocabulary Object Detection (OVOD).
Under this OVOD setting, the detector is trained with only base classes, then tested on both base and novel classes.
In ViLD, an object vocabulary (see Figure. \ref{fig:teaser}) is given to obtain text embeddings (a.k.a. class embeddings) through pretrained text encoder.
% to replace the classifier weights.
The object class is predicted by finding the best matched one from the embedded object vocabulary, and the bounding box is obtained from the class-agnostic regression head.

From recent works on OVOD \cite{gu2022openvocabulary,zhou2022detecting}, it has been widely studied that the proposal generator in a trained detector can generalize to novel classes well even the detector is trained only on the base classes.
However, the current performance of open-vocabulary object detector degenerates and results in low AP when generalized to novel classes.
Intuitively, the low performance is caused by the uncertain matching between proposal features and class embeddings.
Under the OVOD setting with a large amount of classes, as shown in Figure \ref{fig:teaser}, many classes (\eg, puffin v.s. bird) are very similar, which may lead to mismatching between proposal features and class embeddings.
Furthermore, as the detector can only be learned on the base classes, the detector fails to recognize unseen novel classes.
There can be a set of highly similar candidate classes (including both base and novel classes) when the detector meets a novel class object, which brings high uncertainty to novel object class prediction.
This infers too close clustering of data points from similar classes in the latent feature space, and thus the decision boundary crosses high density regions \cite{zhou2003learning,grandvalet2004semi}.
Meanwhile, the trained detector may misclassify a novel class object as background \cite{joseph2021towards,gupta2022ow}.
Current OVOD methods can not handle these issues effectively.
DetPro \cite{du2022learning} makes class embeddings learnable, while other works \cite{zhou2022detecting, feng2022promptdet, bangalath2022bridging} train the model with more classes.
Since only the base classes are given during training, the corresponding class embeddings in DetPro \cite{du2022learning} result in overfitting on the base classes, which has been pointed out in few-shot vision-language learning \cite{zhou2022cocoop}.
In \cite{zhou2022detecting, feng2022promptdet, bangalath2022bridging}, they supplement lots of extra data from ImageNet21k \cite{deng2009imagenet}, Conceptual Captions \cite{sharma2018conceptual}, or LAION-400M \cite{schuhmann2021laion} in the training process, but it is unfair by including novel classes during training as demonstrated in OWL-ViT \cite{minderer2022simple}.

To alleviate the overfitting issue more efficiently, we propose a meta prompt learning scheme. Specifically, during each training iteration, in addition to batch-wise training data and its annotations, we randomly sample more class names from base object vocabulary to simulate the novel-class-emerging scenario as in OVOD setting. 
Then, these enriched text samples are used to learn prompts of text representations. 
With meta representation sampling scheme, we can obtain more discriminative text representations and thus improve the generalization ability to novel classes.
Meanwhile, instead of randomly initializing the background class embedding, we learn the background prompt representation. 
With the learnable background class embedding, the detector can better distinguish the negative (background) proposals from positive (foreground) ones.
The object vocabulary list can be extended with more easily-accessed class names and to further alleviate the overfitting issue (see Table \ref{tab:lvis}).
Further, we incorporate an instance-level contrastive learning scheme to promote the intra-class compactness and inter-class separation, which expands low-density regions in the latent feature space by narrowing the cluster of base classes during the detector training.
With such instance contrastive learning strategy, the novel classes are potentially separated from the base ones in the latent space and thus benefits novel class generalization.
As shown in Figure \ref{fig:novel_error}, MIC decreases the error rate of predicting novel objects as base ones significantly, from 50.4\% to 19.7\%. These results demonstrate the effectiveness of our method for alleviating the extreme overfitting issue.

Our main contributions are summarized as:
1) We introduce a novel meta prompt learning scheme to simulate a novel-class-emerging scenario, which boosts the generalization ability. Meanwhile, the learnable background prompt is incorporated to help the detector distinguish the positive and negative proposals.
2) We propose an instance-level contrastive learning strategy to promote intra-class compactness and inter-class separation, in which the contrastive pairs are built among foreground and background proposal samples. 
3) We conduct extensive experiments on the benchmark dataset LVIS. Without knowledge distillation, ensemble model or extra training data, our method outperforms previous SOTA methods equipped with these complex techniques. Most importantly, our method shows great generalization ability on directly transferring to other datasets, such as COCO and Objects365.

\section{Related Work}
\paragraph{Open-Vocabulary Object Detection.}
Classical object detectors \cite{ren2015faster,he2017mask,lin2017focal,tian2019fcos, carion2020end,zhu2021deformable,meng2021conditional} heavily rely on the large-scale training data and can not generalize to unseen classes during inference.
To alleviate these issues, lots of specific methods have been designed, such as semi/self-supervised \cite{tang2021proposal,su2022self}, few/zero-shot \cite{sun2021fsce,huang2022robust}, and open-world detection \cite{joseph2021towards,han2022expanding}.
However, zero-shot detection can only tackle in-domain
setting, such as splitting COCO as seen and unseen classes, while open-world detection can only detect the unknown objects without classifying them.
Recently, ViLD \cite{gu2022openvocabulary} proposes a new setting called Open-Vocabulary Object Detection (OVOD), which aims to detect any objects within an object vocabulary in an input image.
To achieve this, ViLD replaces the traditional classifier weights with the class embeddings generated by CLIP \cite{radford2021learning} to make it generalize to novel classes.
Following ViLD, some works are done to improve the detection performance by learning prompt representations \cite{du2022learning} or supplementing extra training data \cite{zhou2022detecting, feng2022promptdet, bangalath2022bridging}.
For the former, the learned prompt representation can be easily overfitting on the base classes, which can be harmful to the generalization ability of the detector.
For the latter, it is unfair to train on the novel classes from extra data, which is not strictly OVOD setting \cite{minderer2022simple}.

\paragraph{Prompt Learning.}
Prompt learning is a light-weight framework to adapt large vision-language pretrained models\cite{radford2021learning,jia2021scaling,li2022grounded} to downstream task.
CoOp \cite{zhou2021learning} proposes to learn the prompt representations but not just use a human-designed prompt template.
CoCoOp \cite{zhou2022cocoop} further handles the issue in CoOp that the model overfits on the base classes during training and fails on the novel classes during inference.
Naively transferring prompt learning to OVOD can also cause overfitting, but the instance-level adaption strategy in CoCoOp is impossible for OVOD with the limited GPU memory.

\begin{figure*}[t]
\centering
\includegraphics[width=\textwidth]{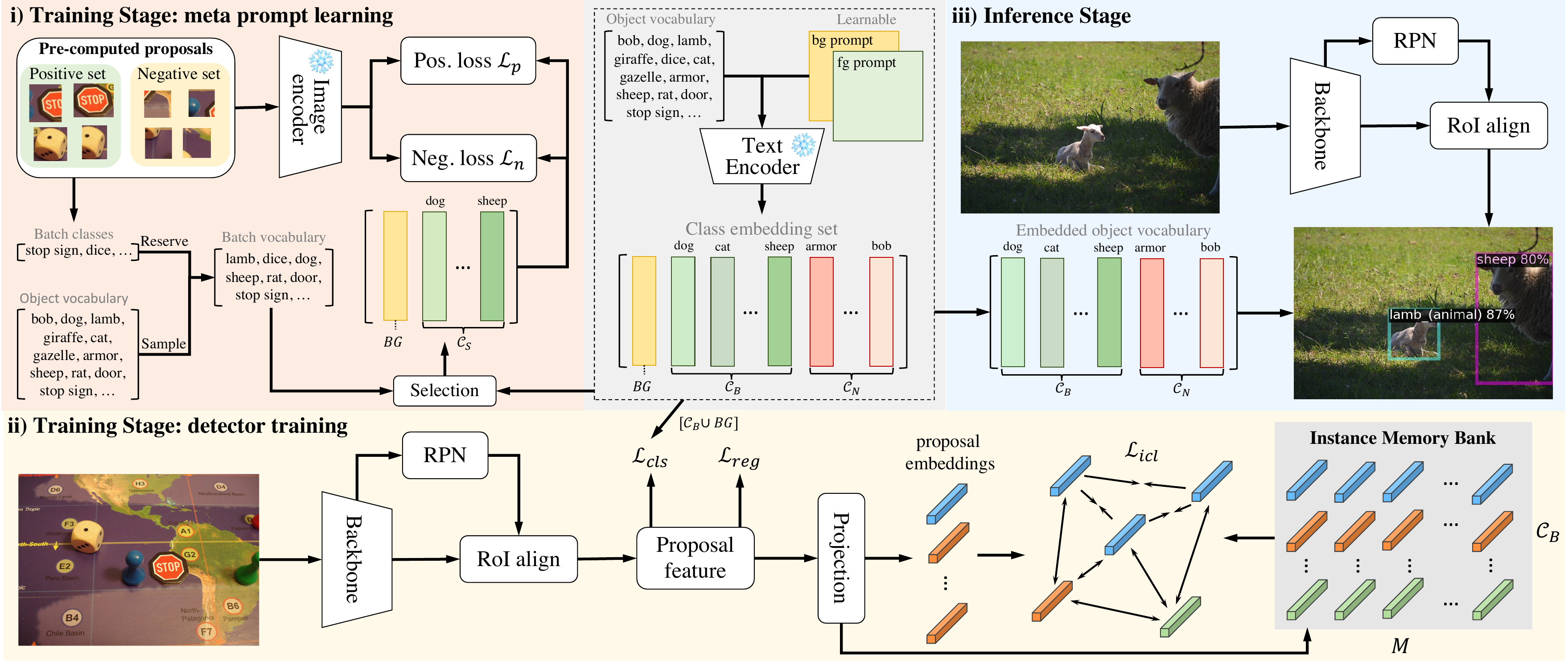}
\vspace{-2mm}
\caption{
{\bf Overview of our proposed method.} The training stage is divided into two consecutive parts: i) meta prompt learning and ii) detector training. 
During i) meta prompt learning, to simulate a novel-class-emerging scenario, we sample a batch-wise varying object vocabulary with $\mathcal{C}_S$ from $\mathcal{C}_B$, which improves the generalization ability of learned foreground prompt. 
Also, we integrate the learnable background prompt to help the model distinguish foreground and background proposals. 
Further, in ii) detector training, we introduce an instance-level contrastive learning scheme to promote intra-class compactness and inter-class separation.
During iii) inference stage, we use the learned foreground prompt representation to generate class embeddings for novel classes.
}
\label{fig:framework}
\end{figure*}

\paragraph{Contrastive Learning.}
Contrastive pretraining \cite{he2020momentum,chen2020simple,grill2020bootstrap,caron2020unsupervised} has been validated effective for learning discriminative feature representations.
Such pretraining strategy is based on contrastive image pairs built from two different strong augmentations of an image.
Further, to increase the diversity of contrasive image pairs, supervised contrastive learning \cite{khosla2020supervised} is proposed.
It builds contrastive pairs across different images, which better exploit the inherent information from images.
In this work, we take the superiority and extend it to instance-level contrastive learning, which promote intra-class compactness and inter-class separation.

\section{Method}

\subsection{Preliminary}
\label{sec:preliminary}
In OVOD, the model can only access base classes $\mathcal{C}_B$ during training and is expected to detect the objects within a wider class vocabulary $\mathcal{C}\!=\!\mathcal{C}_B\cup\mathcal{C}_N$, where $\mathcal{C}_N$ represents the set of novel classes.
We adopt CLIP \cite{radford2021learning} as our vision-language model during meta prompt learning, in which the text encoder $E_\mathcal{T}(\cdot)$ produces the class embeddings and the image encoder $E_\mathcal{I}(\cdot)$ outputs the image embeddings.
ViLD \cite{gu2022openvocabulary} takes the class embeddings obtained from prompt engineering, which is not effective enough for OVOD since the text encoder $E_\mathcal{T}(\cdot)$ is highly sensitive to prompt templates.
Regarding this issue, DetPro \cite{du2022learning} learns a class-shared foreground prompt representation following CoOp \cite{zhou2021learning} instead of using human-defined prompt templates.
The prompt representation $\boldsymbol{V}_{c}$ for class $c \in \mathcal{C}_B$ is $\boldsymbol{V}_{c}=\left\{\boldsymbol{v}_{1}, \boldsymbol{v}_{2}, \dots, \boldsymbol{v}_{L_p}, \boldsymbol{w}_{c}\right\}$,
where $\boldsymbol{v}_{i}$ denotes the $i$-th learnable context vector, $L_p$ is the length of learnable context, and $\boldsymbol{w}_{c}$ is the fixed word embedding for class $c$.
Here, context vectors can be analogue to the human-designed prompt, such as ``a photo of''.
The word embedding $\boldsymbol{w}_{c}$ is obtained from the corresponding class name, and the learnable context vectors $\boldsymbol{V}_{fg}=\left\{\boldsymbol{v}_{i}\right\}_{i=1}^{L_p}$ are randomly initialized.
The class embedding $\boldsymbol{t}_{c}$ of class $c$ is obtained by $\boldsymbol{t}_{c}=E_\mathcal{T}\left(\boldsymbol{V}_{c}\right)$.
Given a positive (foreground) proposal, it is fed into image encoder $E_\mathcal{I}(\cdot)$ to obtain its embedding $\boldsymbol{f}_p$, 
and the probability of $\boldsymbol{f}_p$ to be classified as class $c$ is $p^p_c=\frac{\exp \left(\operatorname{sim}\left(\boldsymbol{f}_p, \boldsymbol{t}_{c}\right) / \tau\right)}{\sum_{i \in \mathcal{C}_{B}} \exp \left(\operatorname{sim}\left(\boldsymbol{f}_p, \boldsymbol{t}_{i}\right) / \tau\right)}$,
where $\tau$ is a temperature parameter, and $\operatorname{sim}\left(\cdot,\cdot\right)$ denotes the cosine similarity.
The cross-entropy loss is used to optimize the context vectors $\boldsymbol{V}_{fg}$ while the base CLIP is fixed:
\begin{equation}
    \mathcal{L}_{p}=-\log p^p_c.
    \label{equ:pos_loss}
\end{equation}
Negative (background) proposals can not be recognized as any foreground objects.
Given a negative proposal, its embedding $\boldsymbol{f}_n$ should be dissimilar to any foreground class embeddings.
In practice, with a large $|\mathcal{C}_B|$, simply optimizing prediction probability $p^n_c$ of a negative proposal to $\frac{1}{|\mathcal{C}_B|}$ can achieve this goal.
Thus, the negative loss is $\mathcal{L}_{\mathrm{n}}=- \frac{1}{\left|\mathcal{C}_{B}\right|} \sum_{c=1}^{\left|\mathcal{C}_{B}\right|} \log p^n_c$, where $p_c^n$ is computed in the same way as $p_c^p$ by replacing $\boldsymbol{f}_p$ with $\boldsymbol{f}_n$.

\subsection{Meta Prompt Learning}
\label{sec:mpl}
With learnable foreground prompt $\boldsymbol{V}_{fg}$, the model performs better compared with prompt engineering.
However, one urging issue is that the learned context can be easily overfitting on base classes $\mathcal{C}_B$ and lack of generalization ability to novel classes $\mathcal{C}_N$.
In this section, we introduce a Meta Prompt Learning (MPL) scheme to promote generalization, as shown in Figure~\ref{fig:framework}.

\paragraph{Meta Representation Sampling.}
Recall that there are $|\mathcal{C}_N|$ novel classes during inference in OVOD.
Regarding this, we aim to simulate such a novel-class-emerging scenario to promote the prompt representation generalizing to novel classes well, in which a batch-wise varying vocabulary $\mathcal{C}_S$ is sampled from the base class vocabulary $\mathcal{C}_B$.
Specifically, given a batch of proposal samples during prompt learning, with the base class embeddings $\boldsymbol{T}_B=\left\{\boldsymbol{t}_{i}\right\}_{i\in \mathcal{C}_B}$, we sample a subset $\boldsymbol{T}_S\!\!=\!\!\left\{\boldsymbol{t}_{i}\right\}_{i\in \mathcal{C}_S}$ of $\boldsymbol{T}_B$ ($\boldsymbol{T}_S\!\!\subset\!\!\boldsymbol{T}_B$).
In this subset $\boldsymbol{T}_S$, the class embeddings of classes existed in the current batch proposal samples are reserved, and other class embeddings are randomly sampled from the remaining base classes.
Then the probability of $\boldsymbol{f}_p$ to be classified as the corresponding class $c$ is changed to
\begin{equation}
    p^p_c=\frac{\exp \left(\operatorname{sim}\left(\boldsymbol{f}_p, \boldsymbol{t}_{c}\right) / \tau\right)}{\sum_{i \in \mathcal{C}_{S}} \exp \left(\operatorname{sim}\left(\boldsymbol{f}_p, \boldsymbol{t}_{i}\right) / \tau\right)}.
    \label{equ:meta_pos_prob}
\end{equation}
With batch-wise varying class embeddings, the learned $\boldsymbol{V}_{fg}$ is more generalizable and robust to the unseen novel classes, which helps the generalization of detector to novel classes.

\paragraph{Background Prompt Representation.}
A typical challenge for a classical detector is how to distinguish the negative (background) proposals from positive (foreground) ones, which becomes more serious in OVOD.
With only base classes during training, detector will easily misclassify novel class objects as background.
Previous works \cite{du2022learning, zhou2022detecting} randomly initialize the background class embedding, while the randomness of the background class embedding can mislead the judgement of detector. 
To this end, we introduce the learnable background prompt representation with $L_n$ learnable context vectors: $\boldsymbol{V}_{bg}=\{\boldsymbol{v}_{1}^{bg}, \boldsymbol{v}_{2}^{bg}, \dots, \boldsymbol{v}_{L_n}^{bg}\}$.
Also, we can obtain the background class embedding $\boldsymbol{t}_{bg}=E_\mathcal{T}\left(\boldsymbol{V}_{bg}\right)$.
Although the background proposals can not be recognized as the foreground objects, sometimes objects are partially located in the background proposals.
So directly including the background proposals into prompt learning can be harmful.
Instead, we consider the background prompt as a negative item to foreground classes.
Then, Eq.~\eqref{equ:meta_pos_prob} becomes
\begin{equation}
    p^p_c=\frac{\exp \left(\operatorname{sim}\left(\boldsymbol{f}_p, \boldsymbol{t}_{c}\right) / \tau\right)}{\sum\limits_{i \in \mathcal{C}_{S}}\!\! \exp \left(\operatorname{sim}\left(\boldsymbol{f}_p, \boldsymbol{t}_{i}\right) / \tau\right) + \exp \left(\operatorname{sim}\left(\boldsymbol{f}_p, \boldsymbol{t}_{bg}\right) / \tau\right)}.
    \label{equ:meta_pos_prob_bg}
\end{equation}
We use $\mathcal{L}_p$ in Eq.~\eqref{equ:pos_loss} to optimize $p^p_c$ in Eq.~\eqref{equ:meta_pos_prob_bg}. For the negative proposals, the prediction probability is computed by 
\begin{equation}
    p^n_c=\frac{\exp \left(\operatorname{sim}\left(\boldsymbol{f}_n, \boldsymbol{t}_{c}\right) / \tau\right)}{\sum_{i \in \mathcal{C}_{S}} \exp \left(\operatorname{sim}\left(\boldsymbol{f}_n, \boldsymbol{t}_{i}\right) / \tau\right)},
    \label{equ:meta_neg_prob}
\end{equation}
and we use the following loss to optimize $p^n_c$:
\begin{equation}
    \mathcal{L}_{\mathrm{n}}=- \frac{1}{\left|\mathcal{C}_{S}\right|} \sum_{c=1}^{\left|\mathcal{C}_{S}\right|} \log p^n_c.
    \label{equ:meta_neg_loss}
\end{equation}

\subsection{Instance Contrastive Learning}
\label{sec:icl}
With $\boldsymbol{V}_{fg}$ and $\boldsymbol{V}_{bg}$ learned in MPL, we then train our detector as shown in Figure \ref{fig:framework}.
As the novel classes are not seen during training, the detector can easily classify a novel class object as a base class one.
Such phenomenon reflects the decision boundary crosses the high density regions by mistake \cite{wang2022c2am}.
In this section, we propose Instance Contrastive Learning (ICL), aiming to expand low-density regions in the latent space by narrowing the cluster of base classes, which potentially pulls the cluster of novel classes from base ones.

\paragraph{Class-balanced Memory Bank.}
Given a large object class vocabulary, previous contrastive learning methods \cite{chen2020simple,he2020momentum} use a huge batch size (\eg, $8192$) to learn with more samples, which is unaffordable for detection.
Even we perform instance-level contrastive learning, the number of proposals contained in a single batch is only around $400$.
So regarding this issue, we build an instance memory bank $\mathcal{Q}$ to collect diverse proposal samples.
Moreover, to avoid frequent class dominating, we make this memory bank class balanced, in which the memory bank $\mathcal{Q}_c$ for each base class $c \in \mathcal{C}_B$ and background class contains $M$ proposal samples.
The memory bank is updated every iteration as the following: 
i) We filter out the foreground proposals with high Intersection of Union (IoU) $>U_{p}$ and identify background proposals with low IoU $<U_{n}$. Here, we set large $U_{p}$ and small $U_{n}$ to make the proposal samples representative.
ii) We sample $m$ ($m<M$) proposal samples that are the most dissimilar with the existing ones in $\mathcal{Q}$ to enrich the diversity of the samples in $\mathcal{Q}$.
Along with the learning process, the memory bank $\mathcal{Q}$ is maintained in a first in and first out manner.

\paragraph{Optimization.}
Directly applying compact regularization to the high dimension proposal embeddings may result in an over-constraint to the network that hinders its convergence. 
Therefore, we introduce a projection network $\mathcal{Z}_\phi$ to map the proposal embeddings $\boldsymbol{f}$ into another low dimensional space as  $\boldsymbol{z}$.
Inspired by supervised contrastive learning \cite{khosla2020supervised}, we propose an instance-level contrastive loss to learn compact embeddings of proposal samples.
The contrastive loss is defined as
\begin{equation}
    \mathcal{L}_{icl}=\frac{1}{N} \sum_{i=1}^{N} \frac{1}{\left|\mathcal{Q}_{\boldsymbol{c}(i)}\right|} \!\!\sum_{j}^{\left|\mathcal{Q}_{\boldsymbol{c}(i)}\right|}\! \log \frac{\exp \left(\boldsymbol{z}_{i}\cdot \boldsymbol{z}_{j} / \gamma\right)}{\sum_{k}^{\left|\mathcal{A}_{\boldsymbol{c}(i)}\right|} \exp \left(\boldsymbol{z}_{i}\cdot \boldsymbol{z}_{k} / \gamma\right)},
    \label{equ:icl}
\end{equation}
where $N$ is the number of proposal samples, $\boldsymbol{c}(i)$ is the class label of $i$-th proposal sample, $\mathcal{Q}_{\boldsymbol{c}(i)}$ denotes the memory bank of class $\boldsymbol{c}(i)$, $\gamma$ is a temperature hyperparameter, and $\mathcal{A}_{\boldsymbol{c}(i)}=\mathcal{Q} \backslash \mathcal{Q}_{\boldsymbol{c}(i)}$.
By expanding the low density regions of base classes cluster in the hidden space, the detector can learn more robust and generalizable feature representations.

\subsection{Training}
\label{sec:training}
The training process of the proposed method consists of two parts: i) meta prompt learning and ii) detector training.
For meta prompt learning, the overall loss is $\mathcal{L}_{mpl} = \mathcal{L}_{p} + \mathcal{L}_{n}$, where $\mathcal{L}_{p}$ and $\mathcal{L}_{n}$ are defined in Eq.~\eqref{equ:pos_loss} and Eq.~\eqref{equ:meta_neg_loss}, respectively.
For detector training, the overall loss is $\mathcal{L}_{det} = \mathcal{L}_{rpn} + \mathcal{L}_{cls} + \mathcal{L}_{reg} + \alpha \mathcal{L}_{icl}$, where $\mathcal{L}_{rpn}$ is RPN loss, $\mathcal{L}_{reg}$ is regression $L1$ loss, $\mathcal{L}_{cls}$ is classification cross-entropy loss, $\mathcal{L}_{icl}$ is defined in Eq.~\eqref{equ:icl}, and $\mathcal{L}_{icl}$ is weighted by $\alpha$.
The overall training algorithm is given in the appendix.

\section{Experiment}

\subsection{Datasets and Evaluation Metrics}
\paragraph{Datasets.} We evaluate our method on the large-scale open-vocabulary benchmark LVIS \cite{gupta2019lvis}.
LVIS v1 is a large-scale dataset with $1203$ categories for object detection and instance segmentation task. 
The long-tailed distribution of LVIS dataset is very suitable for OVOD setting. 
We take frequent and common classes as the base classes $\mathcal{C}_B$ ($866$ classes), and rare classes as the novel classes $\mathcal{C}_N$ ($337$ classes). This dataset contains $100$k and $20$k images for training and validation.
The models are trained only on base classes and evaluated on both base and novel classes.
We also conduct transfer experiments to validate the generalization ability by directly evaluating LVIS-trained model on Pascal VOC \cite{everingham2010pascal}, COCO \cite{lin2014microsoft}, and Objects365 \cite{shao2019objects365}.
The implementation details and training time comparison are described in the appendix.

\paragraph{Evaluation Metrics.} 
Following the previous works \cite{gu2022openvocabulary,du2022learning,zhou2022detecting}, we use Average Precision (AP) to evaluate the performance of our model.
% for a fair comparison.
For LVIS, we take $\mathrm{AP}_r$ as the main metric as it is the performance of model generalizing to the novel classes $\mathcal{C}_N$, and also report $\mathrm{AP}_c$, $\mathrm{AP}_f$, and $\mathrm{AP}$.
For transfer experiments, we report $\mathrm{AP}, \mathrm{AP}_{50}, \mathrm{AP}_{75}, \mathrm{AP}_{s}, \mathrm{AP}_{m}$ and $\mathrm{AP}_{l}$.

\begin{table}[t]
  \centering

\scalebox{0.7}{

\begin{tabular}{l|ccc|cccc|cccc}
\toprule
\multirow{2}[2]{*}{Method} & \multirow{2}[2]{*}{KD?} & \multirow{2}[2]{*}{Ens?} & \multirow{2}[2]{*}{Extra data?} & \multicolumn{4}{c|}{Detection} & \multicolumn{4}{c}{Instance segmentation} \\
      &       &       &       & $\mathrm{AP}_r$ & \textcolor[rgb]{ .651,  .651,  .651}{$\mathrm{AP}_c$} & \textcolor[rgb]{ .651,  .651,  .651}{$\mathrm{AP}_f$} & \textcolor[rgb]{ .651,  .651,  .651}{$\mathrm{AP}$} & $\mathrm{AP}_r$ & \textcolor[rgb]{ .651,  .651,  .651}{$\mathrm{AP}_c$} & \textcolor[rgb]{ .651,  .651,  .651}{$\mathrm{AP}_f$} & \textcolor[rgb]{ .651,  .651,  .651}{$\mathrm{AP}$} \\
\midrule
ViLD \cite{gu2022openvocabulary} & yes   & yes   & no    & 16.7  & \textcolor[rgb]{ .651,  .651,  .651}{26.5 } & \textcolor[rgb]{ .651,  .651,  .651}{34.2 } & \textcolor[rgb]{ .651,  .651,  .651}{27.8 } & 16.6  & \textcolor[rgb]{ .651,  .651,  .651}{24.6 } & \textcolor[rgb]{ .651,  .651,  .651}{30.3 } & \textcolor[rgb]{ .651,  .651,  .651}{25.5 } \\
RegionCLIP \cite{zhong2022regionclip} & no    & no    & CC3M  & 17.1  & \textcolor[rgb]{ .651,  .651,  .651}{27.4 } & \textcolor[rgb]{ .651,  .651,  .651}{34.0 } & \textcolor[rgb]{ .651,  .651,  .651}{28.2 } & -     & \textcolor[rgb]{ .651,  .651,  .651}{-} & \textcolor[rgb]{ .651,  .651,  .651}{-} & \textcolor[rgb]{ .651,  .651,  .651}{-} \\
DetPro \cite{du2022learning} & yes   & yes   & no    & 20.8  & \textcolor[rgb]{ .651,  .651,  .651}{27.8 } & \textcolor[rgb]{ .651,  .651,  .651}{32.4 } & \textcolor[rgb]{ .651,  .651,  .651}{28.4 } & 19.8  & \textcolor[rgb]{ .651,  .651,  .651}{25.6 } & \textcolor[rgb]{ .651,  .651,  .651}{28.9 } & \textcolor[rgb]{ .651,  .651,  .651}{25.9 } \\
OV-DETR \cite{zang2022open} & yes   & no    & no    & -     & \textcolor[rgb]{ .651,  .651,  .651}{-} & \textcolor[rgb]{ .651,  .651,  .651}{-} & \textcolor[rgb]{ .651,  .651,  .651}{-} & 17.4  & \textcolor[rgb]{ .651,  .651,  .651}{25.0 } & \textcolor[rgb]{ .651,  .651,  .651}{32.5 } & \textcolor[rgb]{ .651,  .651,  .651}{26.6 } \\
PromptDet \cite{feng2022promptdet} & no    & no    & LAION-400M & -     & \textcolor[rgb]{ .651,  .651,  .651}{-} & \textcolor[rgb]{ .651,  .651,  .651}{-} & \textcolor[rgb]{ .651,  .651,  .651}{-} & 19.0  & \textcolor[rgb]{ .651,  .651,  .651}{18.5 } & \textcolor[rgb]{ .651,  .651,  .651}{25.8 } & \textcolor[rgb]{ .651,  .651,  .651}{21.4 } \\
Detic \cite{zhou2022detecting} & no    & no    & CC3M  & -     & \textcolor[rgb]{ .651,  .651,  .651}{-} & \textcolor[rgb]{ .651,  .651,  .651}{-} & \textcolor[rgb]{ .651,  .651,  .651}{-} & 19.8  & \textcolor[rgb]{ .651,  .651,  .651}{-} & \textcolor[rgb]{ .651,  .651,  .651}{-} & \textcolor[rgb]{ .651,  .651,  .651}{31.0 } \\
Rasheed \etal \cite{bangalath2022bridging} & yes   & no    & ImageNet21k & -     & \textcolor[rgb]{ .651,  .651,  .651}{-} & \textcolor[rgb]{ .651,  .651,  .651}{-} & \textcolor[rgb]{ .651,  .651,  .651}{-} & 19.3  & \textcolor[rgb]{ .651,  .651,  .651}{23.6 } & \textcolor[rgb]{ .651,  .651,  .651}{27.9 } & \textcolor[rgb]{ .651,  .651,  .651}{24.1 } \\
\midrule
\rowcolor[rgb]{ .949,  .949,  .949} MIC (ours) & no    & no    & no    & \textbf{22.1} & \textcolor[rgb]{ .651,  .651,  .651}{33.9 } & \textcolor[rgb]{ .651,  .651,  .651}{40.0 } & \textcolor[rgb]{ .651,  .651,  .651}{33.8} & \textbf{20.3} & \textcolor[rgb]{ .651,  .651,  .651}{30.6 } & \textcolor[rgb]{ .651,  .651,  .651}{35.2 } & \textcolor[rgb]{ .651,  .651,  .651}{30.6 } \\
\rowcolor[rgb]{ .949,  .949,  .949} MIC* (ours) & no    & no    & 100 class names & \textbf{22.9} & \textcolor[rgb]{ .651,  .651,  .651}{34.0} & \textcolor[rgb]{ .651,  .651,  .651}{39.9} & \textcolor[rgb]{ .651,  .651,  .651}{34.4} & \textbf{20.8} & \textcolor[rgb]{ .651,  .651,  .651}{30.5} & \textcolor[rgb]{ .651,  .651,  .651}{35.4} & \textcolor[rgb]{ .651,  .651,  .651}{30.7} \\
\bottomrule
\end{tabular}%

}
\vspace{2mm}
  \caption{{\bf Comparison of our method with previous SOTA methods on LVIS benchmark.} 
  Note: KD (knowledge distillation); Ens (ensemble model). * indicates we train the prompts with 100 extra class names during MPL.
  }
    \label{tab:lvis}%

\end{table}%

\begin{table}[t]
  \centering

  \scalebox{0.7}{

\begin{tabular}{l|cc|cccccc|cccccc}
\toprule
\multirow{2}[2]{*}{Method} & \multicolumn{2}{c|}{Pascal VOC} & \multicolumn{6}{c|}{COCO}                     & \multicolumn{6}{c}{Objects365} \\
      & $\mathrm{AP}_{50}$ & $\mathrm{AP}_{75}$ & $\mathrm{AP}$ & $\mathrm{AP}_{50}$ & $\mathrm{AP}_{75}$ & $\mathrm{AP}_s$ & $\mathrm{AP}_m$ & $\mathrm{AP}_l$ & $\mathrm{AP}$ & $\mathrm{AP}_{50}$ & $\mathrm{AP}_{75}$ & $\mathrm{AP}_s$ & $\mathrm{AP}_m$ & $\mathrm{AP}_l$ \\
\midrule
Supervised & 78.5  & 49.0  & 46.5  & 67.6  & 50.9  & 27.1  & 67.6  & 77.7  & 25.6  & 38.6  & 28.0  & 16.0  & 28.1  & 36.7 \\
\midrule
ViLD \cite{gu2022openvocabulary} & 73.9  & 57.9  & 34.1  & 52.3  & 36.5  & 21.6  & 38.9  & 46.1  & 11.5  & 17.8  & 12.3  & 4.2   & 11.1  & 17.8 \\
DetPro \cite{du2022learning} & \textbf{74.6} & 57.9  & 34.9  & 53.8  & 37.4  & 22.5  & 39.6  & 46.3  & 12.1  & 18.8  & 12.9  & 4.5   & 11.5  & 18.6 \\
\midrule
\rowcolor[rgb]{ .949,  .949,  .949} MIC (ours)  & 73.0  & \textbf{58.3} & \textbf{39.2} & \textbf{56.8} & \textbf{42.2} & \textbf{27.2} & \textbf{43.1} & \textbf{51.1} & \textbf{14.0} & \textbf{20.1} & \textbf{15.2} & \textbf{6.6} & \textbf{16.6} & \textbf{24.6} \\
\bottomrule
\end{tabular}%

    }
\vspace{2mm}

  \caption{{\bf Comparison of our method with previous SOTA methods on transfer experiments.} We directly evaluate LVIS-trained model on Pascal VOC test set, COCO validation set and Objects365 validation set, together with a supervised baseline.}
\label{tab:cross_dataset}%

\end{table}%

\subsection{Main Results}

\paragraph{Experiment on LVIS.} 
We select the most recent SOTA OVOD methods for comparison, including ViLD \cite{gu2022openvocabulary}, RegionCLIP \cite{zhong2022regionclip}, DetPro \cite{du2022learning}, OV-DETR \cite{zang2022open}, PromptDet \cite{feng2022promptdet}, Detic \cite{zhou2022detecting}, and Rasheed \etal \cite{bangalath2022bridging}.
From Table \ref{tab:lvis}, it can be seen that without knowledge distillation and ensemble model, our method outperforms DetPro by $1.3\%$ bbox $\mathrm{AP}_r$ and $0.5\%$ mask $\mathrm{AP}_r$.
Moreover, without extra training data, our method outperforms PromptDet, Detic, and Rasheed \etal \cite{bangalath2022bridging} by $1.3\%$, $0.5\%$, and $1.0\%$ mask $\mathrm{AP}_r$, respectively.
We also train the prompts with extra 100 class names from ImageNet21k and find that bbox $\mathrm{AP}_r$ can be further improved to $22.9\%$.
Such results demonstrate the effectiveness and robustness of our method when generalizing to novel classes.

\begin{table}[t]
\centering
\begin{minipage}{0.56\linewidth}
\centering

\makeatletter\def\@captype{table}\makeatother
  \scalebox{0.7}{

\begin{tabular}{cc|cc|cccc}
\toprule
\multicolumn{2}{c|}{Prompt} & \multicolumn{2}{c|}{Strategy} & \multicolumn{4}{c}{Detection} \\
FG    & BG    & MPL   & ICL   & $\mathrm{AP}_r$ & \textcolor[rgb]{ .651,  .651,  .651}{$\mathrm{AP}_c$} & \textcolor[rgb]{ .651,  .651,  .651}{$\mathrm{AP}_f$} & \textcolor[rgb]{ .651,  .651,  .651}{$\mathrm{AP}$} \\
\midrule
fixed & \XSolidBrush & \XSolidBrush & \XSolidBrush & 17.6  & \textcolor[rgb]{ .651,  .651,  .651}{34.4 } & \textcolor[rgb]{ .651,  .651,  .651}{40.2 } & \textcolor[rgb]{ .651,  .651,  .651}{33.8 } \\
learnable & \XSolidBrush & \XSolidBrush & \XSolidBrush & 19.7  & \textcolor[rgb]{ .651,  .651,  .651}{34.0 } & \textcolor[rgb]{ .651,  .651,  .651}{39.8 } & \textcolor[rgb]{ .651,  .651,  .651}{33.8 } \\
learnable & \XSolidBrush & \Checkmark & \XSolidBrush & 20.6  & \textcolor[rgb]{ .651,  .651,  .651}{33.5} & \textcolor[rgb]{ .651,  .651,  .651}{39.8} & \textcolor[rgb]{ .651,  .651,  .651}{33.7} \\
learnable & learnable & \Checkmark & \XSolidBrush & 21.2  & \textcolor[rgb]{ .651,  .651,  .651}{34.0} & \textcolor[rgb]{ .651,  .651,  .651}{39.9} & \textcolor[rgb]{ .651,  .651,  .651}{34.1} \\
\rowcolor[rgb]{ .949,  .949,  .949} learnable & learnable & \Checkmark & \Checkmark & \textbf{22.1} & \textcolor[rgb]{ .651,  .651,  .651}{33.9} & \textcolor[rgb]{ .651,  .651,  .651}{40.0} & \textcolor[rgb]{ .651,  .651,  .651}{34.2} \\
\bottomrule
\end{tabular}%

}
\vspace{2mm}
\caption{{\bf Effect of different components} of our approach. Note: FG (foreground prompt); BG (background prompt).} 
\label{tab:overallanalysis}
	
 \end{minipage}
\begin{minipage}{0.4\linewidth}
\centering

\subfloat[Context lengths]{\label{tab:context_length}
    \resizebox{0.75\linewidth}{!}{%

\begin{tabular}{l|ccc}
\toprule
$[L_p$, $L_n]$ & [4, 6] & \cellcolor[rgb]{ .949,  .949,  .949} [8, 10] & [16, 18] \\
\midrule
$\mathrm{AP}_r$ & 25.2  & \cellcolor[rgb]{ .949,  .949,  .949} \textbf{26.4} & 25.8 \\
\textcolor[rgb]{ .651,  .651,  .651}{$\mathrm{AP}$} & \textcolor[rgb]{ .651,  .651,  .651}{39.3} & \cellcolor[rgb]{ .949,  .949,  .949} \textcolor[rgb]{ .651,  .651,  .651}{40.1} & \textcolor[rgb]{ .651,  .651,  .651}{39.7} \\
\bottomrule
\end{tabular}%

    }
}
\\
\subfloat[Different positions of class token]{\label{tab:token_position}
    \resizebox{0.75\linewidth}{!}{
\begin{tabular}{l|ccc}
\toprule
Position & Front & Middle & \cellcolor[rgb]{ .949,  .949,  .949} End \\
\midrule
$\mathrm{AP}_r$ & 23.8  & 25.4  & \cellcolor[rgb]{ .949,  .949,  .949} \textbf{26.4} \\
\textcolor[rgb]{ .651,  .651,  .651}{$\mathrm{AP}$} & \textcolor[rgb]{ .651,  .651,  .651}{39.0} & \textcolor[rgb]{ .651,  .651,  .651}{39.8} & \cellcolor[rgb]{ .949,  .949,  .949} \textcolor[rgb]{ .651,  .651,  .651}{40.1} \\
\bottomrule
\end{tabular}%

    }
}
\vspace{2mm}
\caption{{\bf Learnable context study.}}

\end{minipage}

\end{table}

\begin{figure}[t]
    \centering
	\begin{minipage}{0.48\linewidth}
		\centering
		\includegraphics[width=\linewidth]{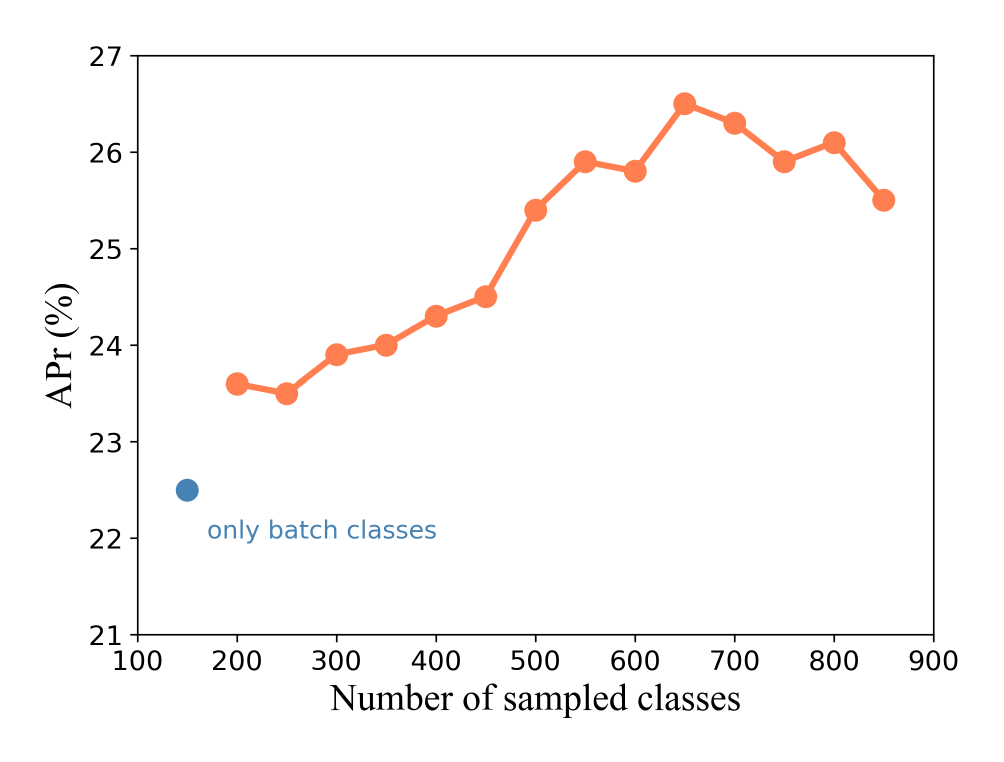}
  \vspace{-2mm}
		\caption{{\bf Sampling strategy in MPL.} We study the effect of sampled  classes.}
\label{fig:mAPr_subclass}
	\end{minipage}
	%\qquad
	\hfill
	\begin{minipage}{0.48\linewidth}
		\centering
       \subfloat[IoU threshold $U_p$ and $U_n$]{\label{tab:icl_ablation_iou}
    \resizebox{\linewidth}{!}{%
\begin{tabular}{l|ccccccccc}
\toprule
$U_n$ & \cellcolor[rgb]{ .949,  .949,  .949} 0.01  & 0.05  & 0.1   & 0.01  & 0.05  & 0.1   & 0.01  & 0.05  & 0.1 \\
$U_p$ & \cellcolor[rgb]{ .949,  .949,  .949} 0.7 & 0.7   & 0.7   & 0.8   & 0.8   & 0.8   & 0.9   & 0.9   & 0.9 \\
\midrule
$\mathrm{AP}_r$ & \cellcolor[rgb]{ .949,  .949,  .949} \textbf{26.4} & 25.1  & 26.0  & 26.0  & 24.7  & 24.3  & 25.4  & 25.2  & 24.8 \\
\textcolor[rgb]{ .651,  .651,  .651}{$\mathrm{AP}$} & \cellcolor[rgb]{ .949,  .949,  .949} \textcolor[rgb]{ .651,  .651,  .651}{40.1} & \textcolor[rgb]{ .651,  .651,  .651}{40.0} & \textcolor[rgb]{ .651,  .651,  .651}{39.8} & \textcolor[rgb]{ .651,  .651,  .651}{39.9} & \textcolor[rgb]{ .651,  .651,  .651}{40.1} & \textcolor[rgb]{ .651,  .651,  .651}{40.2} & \textcolor[rgb]{ .651,  .651,  .651}{39.7} & \textcolor[rgb]{ .651,  .651,  .651}{39.9} & \textcolor[rgb]{ .651,  .651,  .651}{39.4} \\
\bottomrule
\end{tabular}%

    }
}
\\
\vspace{-2mm}
\subfloat[Batch sampling size $m$ and memory size $M$]{\label{tab:icl_ablation_memory}
    \resizebox{\linewidth}{!}{%
\begin{tabular}{l|ccccccccc}
\toprule
$m$   & 8     & 16    & 32    & 8     & 16    & 32    & 8     & \cellcolor[rgb]{ .949,  .949,  .949} 16  & 32  \\
$M$   & 64    & 64    & 64    & 128   & 128   & 128   & 256   & \cellcolor[rgb]{ .949,  .949,  .949} 256  & 256  \\
\midrule
$\mathrm{AP}_r$ & 24.3  & 24.5  & 24.6  & 26.0  & 25.5  & 24.9  & 25.7  & \cellcolor[rgb]{ .949,  .949,  .949} \textbf{26.4} & 23.9 \\
\textcolor[rgb]{ .651,  .651,  .651}{$\mathrm{AP}$} & \textcolor[rgb]{ .651,  .651,  .651}{39.8} & \textcolor[rgb]{ .651,  .651,  .651}{40.2} & \textcolor[rgb]{ .651,  .651,  .651}{39.8} & \textcolor[rgb]{ .651,  .651,  .651}{39.9} & \textcolor[rgb]{ .651,  .651,  .651}{39.7} & \textcolor[rgb]{ .651,  .651,  .651}{40.2} & \textcolor[rgb]{ .651,  .651,  .651}{39.6} & \cellcolor[rgb]{ .949,  .949,  .949} \textcolor[rgb]{ .651,  .651,  .651}{40.1} & \textcolor[rgb]{ .651,  .651,  .651}{39.4} \\
\bottomrule
\end{tabular}%

    }
}
\vspace{2mm}
        \captionof{table}{{\bf Sampling strategy in ICL.} (a) Effect of foreground and background instance IoU threshold $U_p$ and $U_n$. (b) Effect of batch sampling size $m$ and memory size $M$.}
	    \end{minipage}
\end{figure}

\paragraph{Transfer Experiment.}
We further validate the generalization ability of our method by directly evaluating LVIS-trained model on Pascal VOC, COCO, and Objects365.
We use the learned prompt representation and class names of corresponding dataset to generate the class embeddings.
From Table \ref{tab:cross_dataset}, we can observe that our method improves the performance by a large margin, especially on more difficult COCO ($+4.3\% \ \mathrm{AP}$) and Objects365 ($+1.9\% \ \mathrm{AP}$).

\begin{figure}[t]
\centering

\begin{minipage}[t]{0.435\textwidth}
\centering
\includegraphics[width=\textwidth]{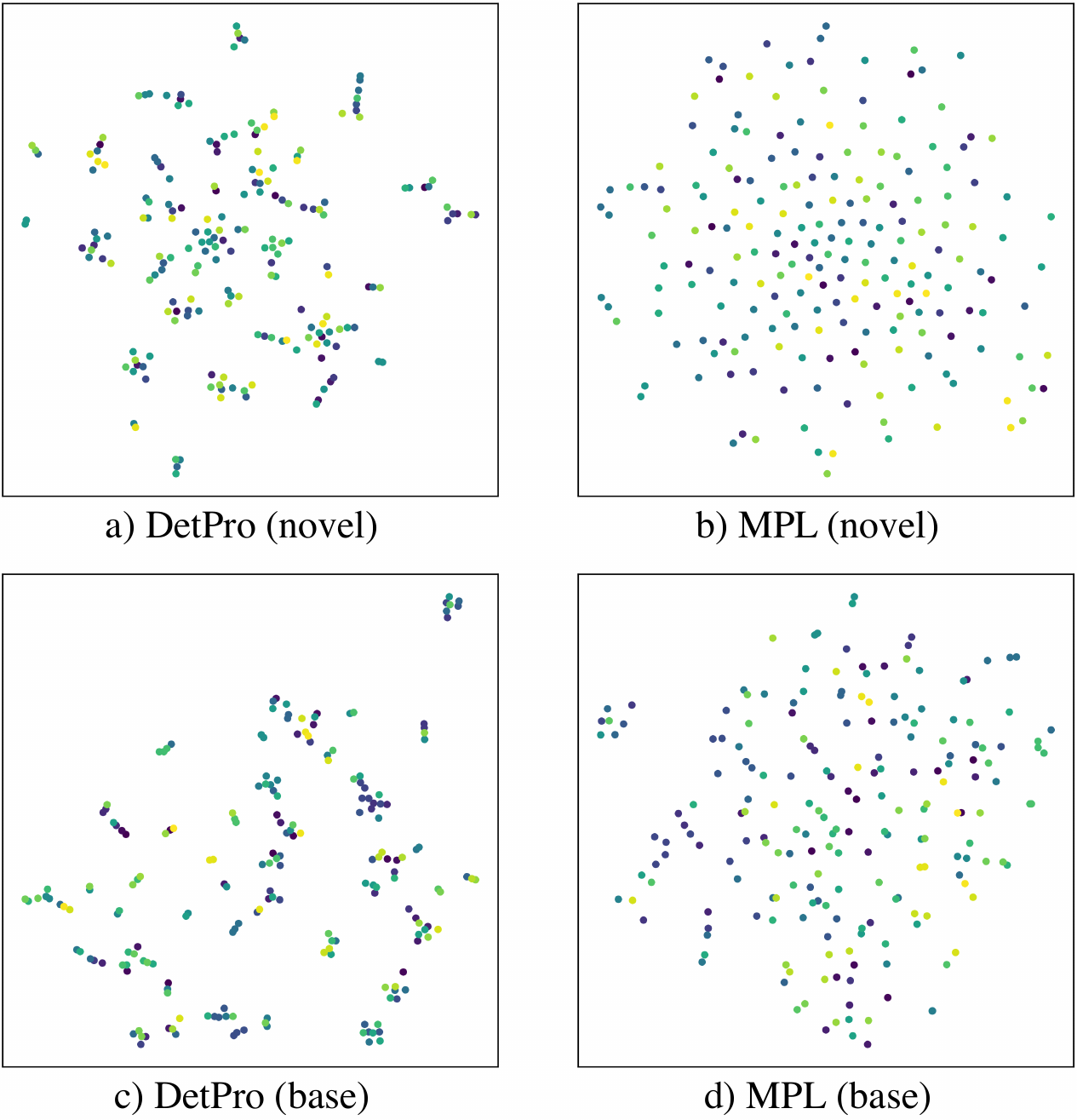}
\vspace{-1mm}
\caption{{\bf t-SNE visualization of class embeddings of LVIS.} We randomly sample 200 novel and base classes from LVIS and use t-SNE to visualize the class embeddings.}
\label{fig:lvis_tsne}
\end{minipage}
\hspace{1mm}
\begin{minipage}[t]{0.525\textwidth}
\centering
\includegraphics[width=\textwidth]{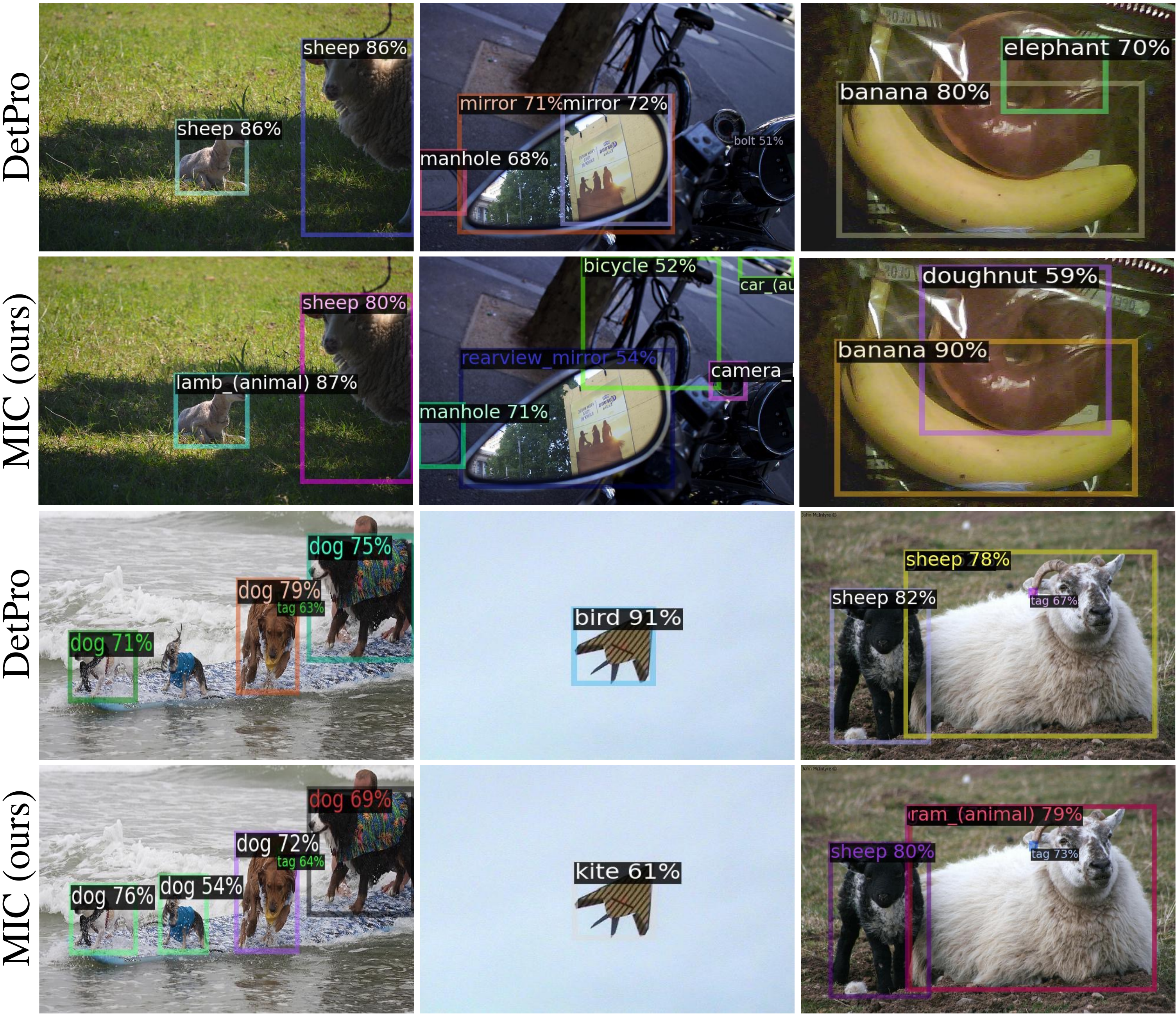}
\vspace{-1mm}
\caption{{\bf Qualitative detection visualization results of our proposed method MIC and DetPro.} Our method could better distinguish similar classes, detect smaller objects, and produce less false positives under diverse complex scenarios.}
\label{fig:detection_visualization}
\end{minipage}
\end{figure}

\subsection{Ablation Studies}
In this subsection, we do comprehensive ablation experiments on LVIS.
To study MPL and ICL schemes, we randomly sample a subset with $5$k images from LVIS validation set for hyper-parameter selection.

\paragraph{Overall Analysis.}
We study the effect of different components in our proposed framework.
As shown in Table \ref{tab:overallanalysis}, MPL helps learnable foreground prompt generalize, with $+0.9\%$ $\mathrm{AP}_r$ improvement.
With the learnable background prompt, the detection performance on novel classes improves by $0.6\%$.
Further, our proposed ICL boosts the performance $\mathrm{AP}_r$ to $22.1\%$, surpassing fixed prompt ($+4.5\% \ \mathrm{AP}_r$) and naive learnable prompt ($+2.4\% \ \mathrm{AP}_r$).

\paragraph{Number of Sampled Classes in MPL.}
We study the model performance when trained with different number of sampled classes in Figure \ref{fig:mAPr_subclass}.
We range the number of classes from $200$ to $850$ in intervals of $50$, along with a special case that only the batch classes are reserved (about $150$ classes).
We can learn that $AP_r$ shows a trend of first rising and then falling as the sample size increases.

\paragraph{Context Lengths.}
We study the effect of using different foreground and background context lengths. 
Due to lack of class word embedding, we set $L_n$ always longer than $L_p$ by $2$.
As shown in Table \ref{tab:context_length}, too short context is not effective while too long context may cause over-fitting.
We set the foreground and background context as $8$ and $10$, respectively.

\paragraph{Position of Class Token.}
In Table \ref{tab:token_position}, we study the effect of different positions of class token in the foreground prompt, including front, middle, and end.
The best position of class token usually depends on the dataset \cite{zhou2021learning}.
We find that inserting it in the end performs best.

\paragraph{Sampling Thresholds in ICL.}
Here, we study the effect of selection thresholds under the different combinations of $U_p$ and $U_n$, as shown in Table \ref{tab:icl_ablation_iou}.
We can learn the following: 1) the model performs better with smaller background selection threshold which infers few overlap between proposal and gt bounding box; 2) the model performs best with foreground selection threshold $0.7$, which demonstrates that too high threshold will filter out too many foreground proposals and lead to sub-optimal solution.

\paragraph{Memory Bank and Sampling Sizes.}
Further, we study the size of memory bank $M$ and batch sampling size $m$.
Intuitively, with more samples in the memory bank, the contrastive regularization can be stronger.
We consider several different combinations of memory size $M$ and batch sampling size $m$, and the results are shown in Table \ref{tab:icl_ablation_memory}. 
We can learn that with memory size $256$ and batch sampling size $16$, the detection achieves the best performance.

\subsection{Visualization Results}
\paragraph{Latent Space Embedding.}
To further validate the effectiveness of our proposed MPL, we randomly sample $200$ novel and base classes from LVIS and use t-SNE \cite{van2008visualizing} to visualize the class embeddings from DetPro and our MPL.
From Figure \ref{fig:lvis_tsne}, it can be seen that MPL can learn more discriminative and generalizable class embeddings, especially for novel classes.

\paragraph{Detection Results.}
We show the superiority of our method by visualizing the detection results on LVIS validation set.
The visualization results are shown in Figure \ref{fig:detection_visualization}.
It can be seen that our method can better distinguish similar categories compared with DetPro, such as sheep and lamb.
Further, our method can detect smaller objects in the complex environments.
Also, with the learnable background prompt, our model produces less false positives.

\section{Conclusion}

This paper proposes a novel framework MIC for open-vocabulary object detection.
MIC consists of two major carefully designed learning schemes, meta prompt and instance contrastive learning.
The proposed meta prompt learning strategy simulates a novel-class-emerging scenario, together with the learnable background prompt representation to help the generalization.
Further, we propose to use instance-level contrastive learning strategy to help expand the low density regions in the latent feature space.
Without complex training techniques and extra training data, extensive experimental results show the strong generalization ability of our proposed method, especially transferring to other datasets, such as COCO and Objects365.

\bigskip
\paragraph{\bf Acknowledgements.} We gratefully acknowledge the support of Mindspore, CANN (Compute Architecture for Neural Networks) and Ascend AI Processor used for this research.

\appendix
% \paragraph{\bf Roadmap.} The appendix is organized as
% follows. Detailed experimental setup are described in Section \ref{appendix:exp}. We further provide more experimental results in Section \ref{appendix:vis}.

\section{Detailed Experimental Setup}
\label{appendix:exp}

Our method is implemented based Detectron2 \cite{wu2019detectron2}.
All the experiments are run on 8 V100 GPUs. The overall algorithm is given in Algorithm \ref{alg:main}.

\paragraph{Transferred Datasets.}
Pascal VOC is a combination of VOC2007 and VOC2012 datasets with $20$ object categories. This dataset contains $5$k images for test.
COCO 2017 is a classical benchmark dataset consisting of $80$ object categories. It contains 5k images for validation.
Objects365 v2 is a object detection dataset with $365$ diverse object classes in the wild. It contains $30$k images for validation.

\paragraph{Meta Prompt Learning.}
The context vectors for foreground and background are both randomly initialized from a Gaussian distribution with $0$ mean and $0.02$ standard deviation. 
And the length of context vectors for foreground and background are set as $8$ and $10$, respectively.
The subset $\boldsymbol{T}_s$ contains $650$ sampled base classes by default.
We use jit version CLIP as vision-language model \cite{radford2021learning} during meta prompt representation learning.
Following DetPro \cite{du2022learning}, we use $10\%$ background proposals and ground-truth foreground proposals are included in the positive proposals for training.
The word embedding for every class is integrated in the end of the learned foreground prompt representation.
The dimension of context vector and word embedding are both $512$.
As demonstrated in DetPro, different positive proposals for the same object can be various, which results in different contexts.
Such difference can be understood as follows: (a) given the ground-truth bounding box of an object, the prompt should be `a photo of'; (b) given a proposal with partial object, the prompt should be `a photo of partial'.
Obviously, the learned prompt representation should be capable to represent these two different templates.
So we train the prompt representation with different level contexts and ensemble the learned prompt representation.
The positive proposals are divided into $5$ levels by IoU range from $0.5$ to $1.0$ with step size $0.1$.
We train the prompt representation for every single level with $\mathcal{L}_p$ in Eq.~(1) and $\mathcal{L}_n$ in Eq.~(5).
The temperature $\tau$ in Eq.~(3) and Eq.~(4) is set as $0.01$.
SGD is used for optimizing the context vectors.
The learning rate is $0.002$ and decayed by a step-wise scheduler, trained for 6 epochs with a batch size $512$.

\begin{algorithm}[t]
\caption{MIC for Open-Vocabulary Object Detection}
\label{alg:main}
\textbf{Input:} fg prompt $\boldsymbol{V}_{fg}$, bg prompt $\boldsymbol{V}_{bg}$, base classes $\mathcal{C}_B$, prompt update steps $K$ and learning rate $\eta_k$, detector $\theta$, training image $I$ with its bbox and class label $(b, c)$, detector training steps $R$ and learning rate $\eta_r$, hyper-parameter $\alpha$\\
% \textbf{Parameter}: Optional list of parameters\\
\textbf{Output:} trained detector $\theta$
\begin{algorithmic}[1] %[1] enables line numbers
\State We start with {\bf procedure A} to learn fg and bg prompts, then the learned prompts are used in {\bf procedure B}.
\Procedure{{\bf A.} Meta Prompt Learning}{}
\For{$k=1\to K$}
\State Sample a batch of precomputed proposals from $\mathcal{C}_B$
\State Forward proposals into CLIP to obtain $\boldsymbol{f}_p$, $\boldsymbol{f}_n$
\State $\boldsymbol{T}_B = \{E_\mathcal{T}(\boldsymbol{V}_{i})\}_{i\in \mathcal{C}_B}$, $\boldsymbol{t}_{bg} = E_\mathcal{T}(\boldsymbol{V}_{bg})$
\State Sample $\boldsymbol{T}_S$ from $\boldsymbol{T}_B$
\Comment{Meta sampling}
\State Compute $p^p_c$ by Eq.~(3), and $p^n_c$ by Eq.~(4)
\State $\mathcal{L}_{p}=-\log p^p_c$
\Comment{Eq.~(1)}
\State $\mathcal{L}_{\mathrm{n}}=- \frac{1}{\left|\mathcal{C}_{S}\right|} \sum_{c=1}^{\left|\mathcal{C}_{S}\right|} \log p^n_c$
\Comment{Eq.~(5)}
\State $\boldsymbol{V}_{fg} \leftarrow \boldsymbol{V}_{fg} - \eta_k\cdot\nabla\mathcal{L}_{p}$ 
\Comment{Update fg prompt}
\State $\boldsymbol{V}_{bg} \leftarrow \boldsymbol{V}_{bg} - \eta_k\cdot\nabla\mathcal{L}_{n}$
\Comment{Update bg prompt}
\EndFor
\EndProcedure

\Procedure{{\bf B.} Detector Training}{}
\For{$r=1\to R$}
\State Sample a batch of data $(I, (b, c))$ from $\mathcal{C}_B$
\State Feed $I$ into detector to obtain $\boldsymbol{f}$ and proposal IoU
\State Filter fg proposal by $U_p$ and bg proposal by $U_n$
\State Update instance memory bank $\mathcal{Q}$
\State Calculate contrastive loss $\mathcal{L}_{icl}$ by Eq.~(6)
\State $\mathcal{L}_{det} = \mathcal{L}_{rpn} + \mathcal{L}_{cls} + \mathcal{L}_{reg} + \alpha \mathcal{L}_{icl}$
\State $\theta \leftarrow \theta - \eta_r\cdot\nabla\mathcal{L}_{det}$ 
\Comment{Update detector $\theta$}
\EndFor
\EndProcedure

\end{algorithmic}
\end{algorithm}

\paragraph{Open-Vocabulary Detector.}
We use ResNet50 \cite{he2016deep} as the backbone and FPN \cite{lin2017feature} architecture.
The default size $M$ of memory bank for each class is $256$ and sampling size $m$ is $16$.
The threshold for selecting proposals $U_p$ and $U_n$ are $0.7$ and $0.01$, respectively.
The temperature $\gamma$ in Eq.~(6) is set as $0.1$ and the weight $\alpha$ of $\mathcal{L}_{icl}$ is initialized as 0.1 and gradually decayed during the detector training to help the convergence of other losses.
Following Detic \cite{zhou2022detecting}, we use CenterNet2 \cite{zhou2021probabilistic} with a ResNet50 backbone \cite{he2016deep} and FPN \cite{lin2017feature} architecture.
The mask prediction head is modified to a class-agnostic one and federated loss \cite{zhou2021probabilistic} is used for training.
Repeat factor sampling \cite{gupta2019lvis} is used to balance long-tailed distributed classes.
Note that CenterNet2 uses a cascade classifier \cite{cai2018cascade}.
The backbone ResNet50 is initialized with the pretrained weights on ImageNet21k \cite{ridnik2021imagenet}.
In ICL, the projection network is a $2$-layer MLP with two linear-relu layers and a normalization layer, and the dimension of proposal feature is projected from $1024$ to $128$.
To improve the diversity of proposal samples in the instance memory bank, we gather and concat all proposal samples from $8$ GPUs before updating the instance memory bank.
We use AdamW \cite{loshchilov2018decoupled} as the optimizer with an initial learning rate $0.0002$ and batch size $64$.
We use EfficientDet style large scale jittering \cite{ghiasi2021simple,tan2020efficientdet}.
To accelerate training, the input images are cropped to $640\times640$ while $800 \times 1333$ for inference.
The detector is trained for a $4\times$ schedule with $90000$ iterations.
We perform $10000$ warmup steps with $0.0001$ warmup factor and use cosine learning schedule \cite{loshchilov2016sgdr}.
To help the convergence of other losses, the weight $\alpha$ of instance-level contrastive loss $\mathcal{L}_{icl}$ is decayed along with the training process by $\alpha = \alpha \times (1 - iter/90000)$.

\section{More Experimental Results}
\label{appendix:vis}
In this section, we show more experimental results, including the comparison of training time, visualization of latent space embedding and detection, which further indicates the effectiveness and robustness of our proposed method.

\paragraph{Training Time.}
We compare the training time of our proposed MIC with previous SOTA two-stage methods, including DetPro \cite{du2022learning}, RegionCLIP \cite{zhong2022regionclip}, PromptDet \cite{feng2022promptdet}, as shown in Table \ref{tab:time}. 
Our MIC is shown to be more efficient than previous two-stage methods.

% Table generated by Excel2LaTeX from sheet '工作表1'
\begin{table}[htbp]
  \centering
\scalebox{0.85}{

    \begin{tabular}{l|cccc}
    \toprule
    Method & RegionCLIP \cite{zhong2022regionclip} & DetPro \cite{du2022learning} & PromptDet \cite{feng2022promptdet} & MIC (ours) \\
    \midrule
    Training Time (GPU hours) & 1064  & 464   & 408   & 368  \\
    \bottomrule
    \end{tabular}%

}
\vspace{2mm}
  \caption{The training time comparison of our method with previous SOTA two-stage methods.}
  \label{tab:time}%
\end{table}%

\begin{figure}[t]
\centering
\includegraphics[width=0.9\linewidth]{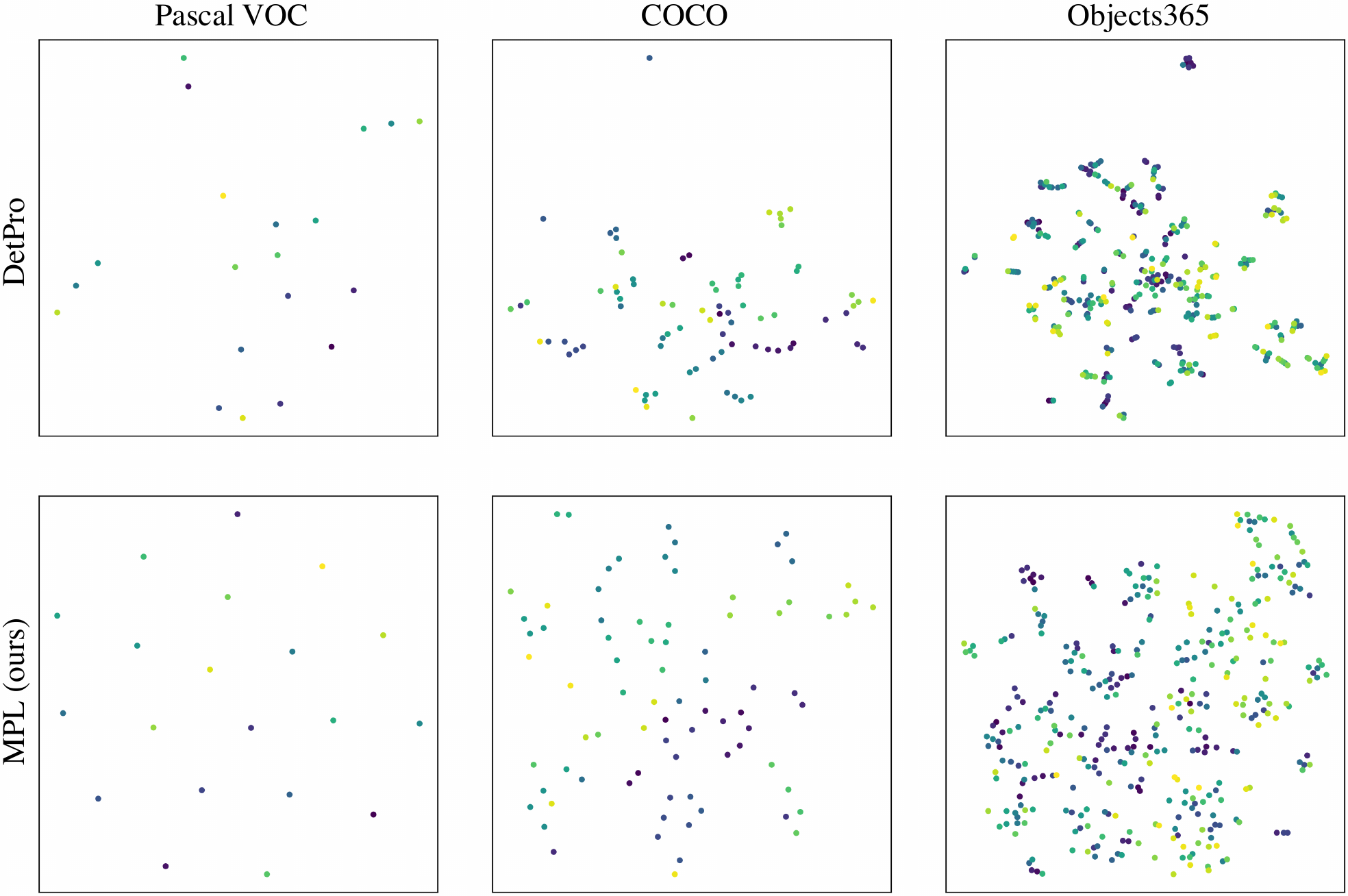}
\vspace{2mm}
\caption{{\bf t-SNE visualization of class embeddings of transferred datasets.} We use t-SNE to visualize the class embeddings of Pascal VOC, COCO, and Objects365 generated from DetPro and our proposed MPL. 
}
\label{fig:transfer_tsne}
\end{figure}

\paragraph{Latent Space Embedding.}
We also use t-SNE to visualize the class embeddings of Pascal VOC, COCO, and Objects365 generated from DetPro and our proposed MPL.
From Figure \ref{fig:transfer_tsne}, we can draw the same conclusion as LVIS.
Under our proposed MPL scheme, the learned prompt representations are more discriminative in the latent space.

% \newpage

\begin{figure}[htbp]
  \centering
  \includegraphics[width=0.6\linewidth]{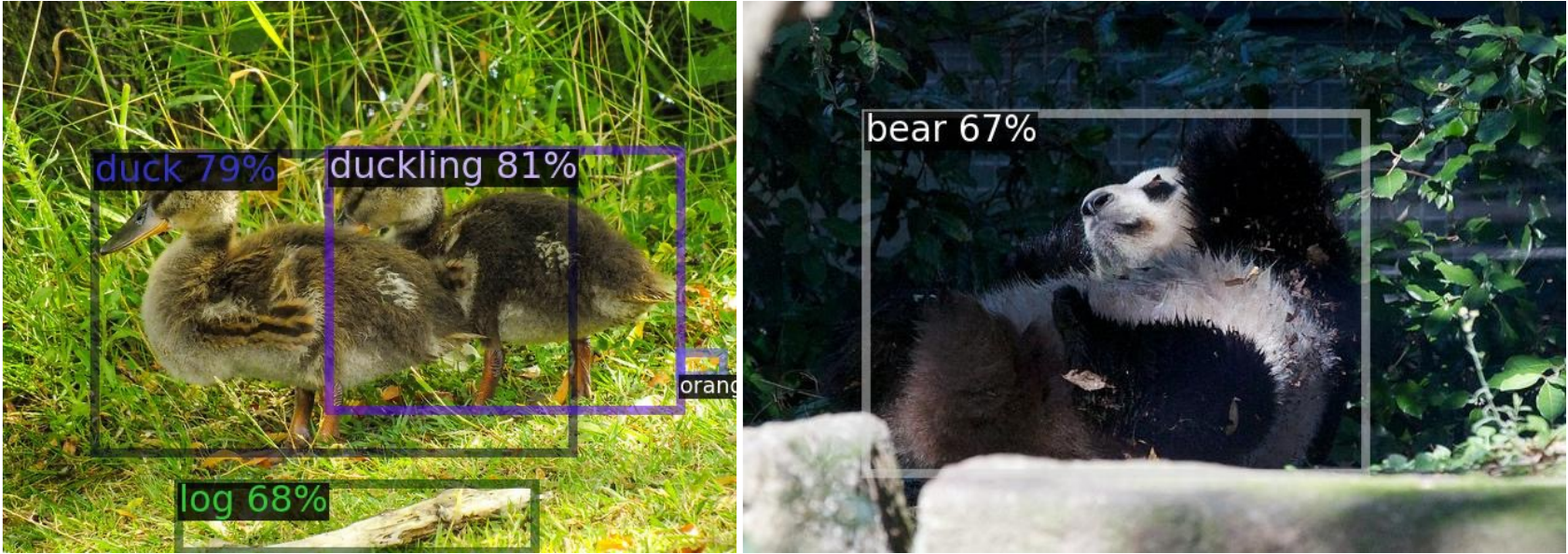}
  \vspace{2mm}
    \caption{Failure cases of MIC.}
   \label{fig:failurecases}
\end{figure}

\paragraph{Study of Failure Cases.}
Although we use meta prompt and instance contrastive learning to improve the discriminative ability of our model, it still suffers from distinguishing some extremely similar classes, such as `duck' and `duckling', and `panda' and `bear' shown in Figure \ref{fig:failurecases}. To better distinguish these categories, we might include outside knowledge base of fine-grained categories in the future work.

\begin{figure}[t]
\centering
\includegraphics[width=\linewidth]{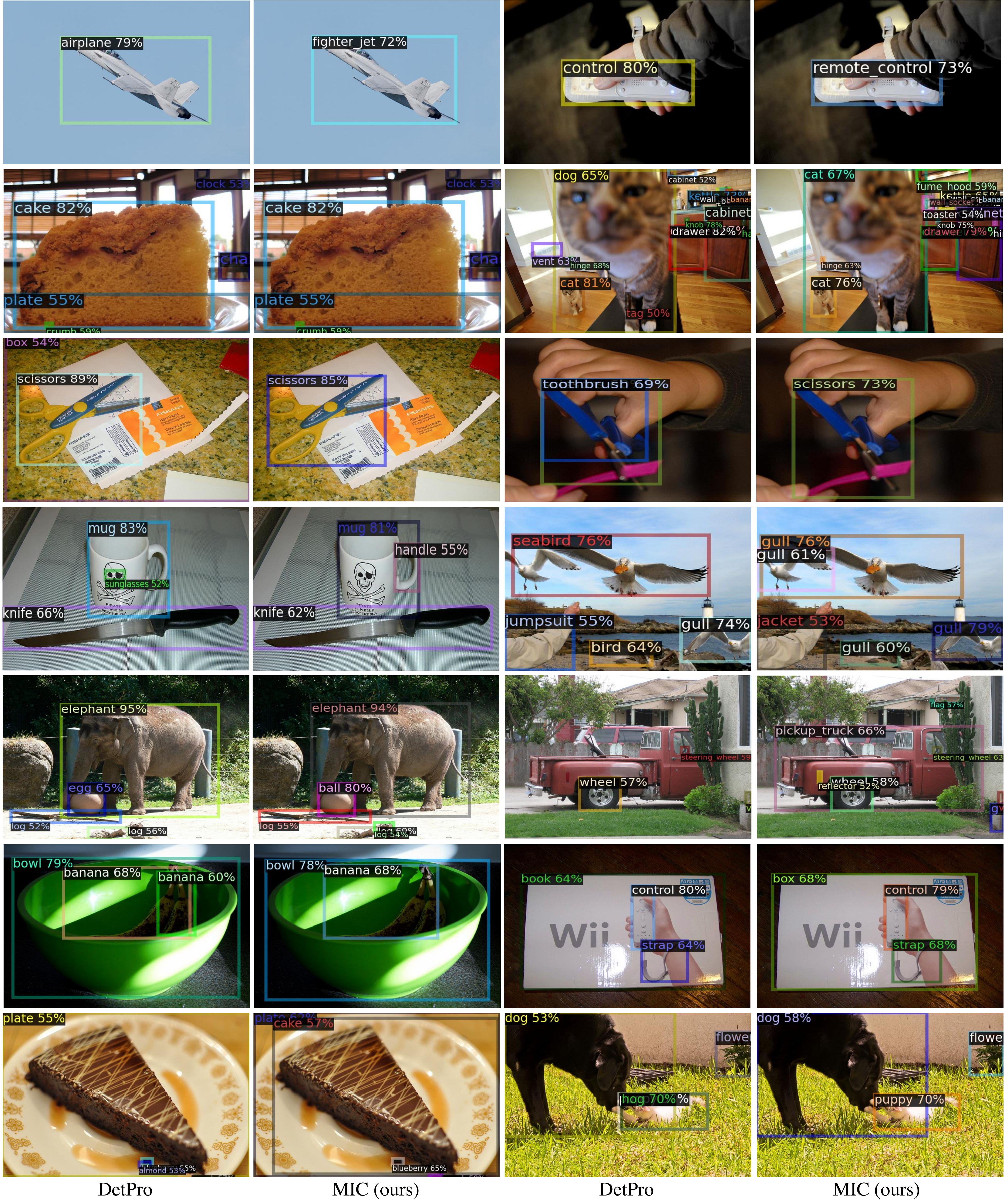}
\vspace{-2mm}
\caption{{\bf More qualitative detection visualization results of our proposed method MIC and DetPro.} Our method could better distinguish similar classes, detect smaller objects, and produce less false positives under diverse scenes.}
\label{fig:detection_visualization_more}
\end{figure}

\paragraph{More Comparisons of Detection Results.}
We show more detection visualization results in Figure \ref{fig:detection_visualization_more}, which further indicates the robustness of our proposed method.

\clearpage

\bibliography{bmvc}

% \clearpage
% \input{sections/appendix}

\end{document}